# Deep Unitary Convolutional Neural Networks

Hao-Yuan Chang[1[0000-0003-2864-7538] (✉)] and Kang L. Wang[1[0000-0002-9363-1279]]

[1] University of California, Los Angeles, Los Angeles CA 90095, USA
{h.chang,klwang}@ucla.edu

**Abstract.** Deep neural networks can suffer from the exploding and vanishing activation problem, in which the networks fail to train properly because the neural signals either amplify or attenuate across the layers and become saturated. While other normalization methods aim to fix the stated problem, most of them have inference speed penalties in those applications that require running averages of the neural activations. Here we extend the unitary framework based on Lie algebra to neural networks of any dimensionalities, overcoming the major constraints of the prior arts that limit synaptic weights to be square matrices. Our proposed unitary convolutional neural networks deliver up to 32% faster inference speeds and up to 50% reduction in permanent hard disk space while maintaining competitive prediction accuracy.

**Keywords:** Neural network, Lie algebra, Image recognition.

## 1 Introduction

### 1.1 Problem Statement

Recent advancements in semiconductor technology [1] have enabled neural networks to grow significantly deeper. This abundant computing power enabled computer scientists to drastically increase the depths of neural networks from the 7-layer LeNet network [2] to the 152-layer contest-wining ResNet architecture [3]. More layers usually lead to higher recognition accuracy because neural networks make decisions by drawing decision boundaries in the high dimensional space [4]. A decision boundary is a demarcation in the feature space that separates the different output classes. The more layers the network has, the more precise these boundaries can be in the high dimensional feature space; thus, they can achieve higher recognition rates [5]. However, deep networks often fail to train properly due to poor convergence.

There are many reasons why a deep network fails to train [6], and the problem that our proposal fixes is the instability of the forward pass, in which neural activations either saturate to infinity or diminish to zero. More precisely, depending on the eigenvalues of the synaptic weight matrices [7], neural signals may grow or attenuate as they travel across neural layers when unbounded activation functions such as the rectified linear units (Relu) are used [8]. The Relu is the most popular nonlinearity due to its computational efficiency. Suppose the activation is extremely large or small; in this case, the weight update will scale proportionally during training, resulting in either a massive or a tiny step.

In short, vanishing and exploding activations occur when the neural signals are not normalized, and the backpropagated gradients either saturate or die out during network training [9]. Although other schemes such as batch normalization [10], learning rate tuning [11], and gradient highways [3] can mitigate the issue, none of these methods eliminate the core problem—the weight matrices have eigenvalues that are larger or smaller than one. Furthermore, most normalization methods have inference time penalties. In this work, we aim to devise a way to fundamentally fix the exploding and vanishing activation problem without slowing down the inference speed.

### 1.2 Proposed Solution

Our proposed solution (Fig. 1) is to eliminate the need to normalize the neural signals after each layer by constraining the weight matrices, $\boldsymbol{W}$, to be unitary. Unitary matrices represent rotations in the n-dimensional space[1]; hence, they preserve the norm (i.e., the amplitude) of the input vector. With this unique property, unitary networks can maintain the neural signal strengths without explicit normalization. This technique allows the designers to eliminate the networks' normalization blocks and make inference faster.

We aim to engineer a way to constrain the weights to be *unitary*. To achieve this, we leverage the previously reported framework for constructing orthogonal matrices in recurrent neural networks using Lie algebra [12], which we will explain briefly in Sect. 2.1. Unlike other approximation methods, this framework guarantees strictly unitary matrices; however, it is currently limited to square matrices. *Our main contribution is that we found a way (Sect. 2.2) to extend the unitary framework based on Lie algebra to weight matrices of any shapes.* By doing so, we expand the applicability of this framework from recurrent neural networks with square weight matrices to any neural network structures, drastically increasing its usefulness in state-of-the-art network architectures.

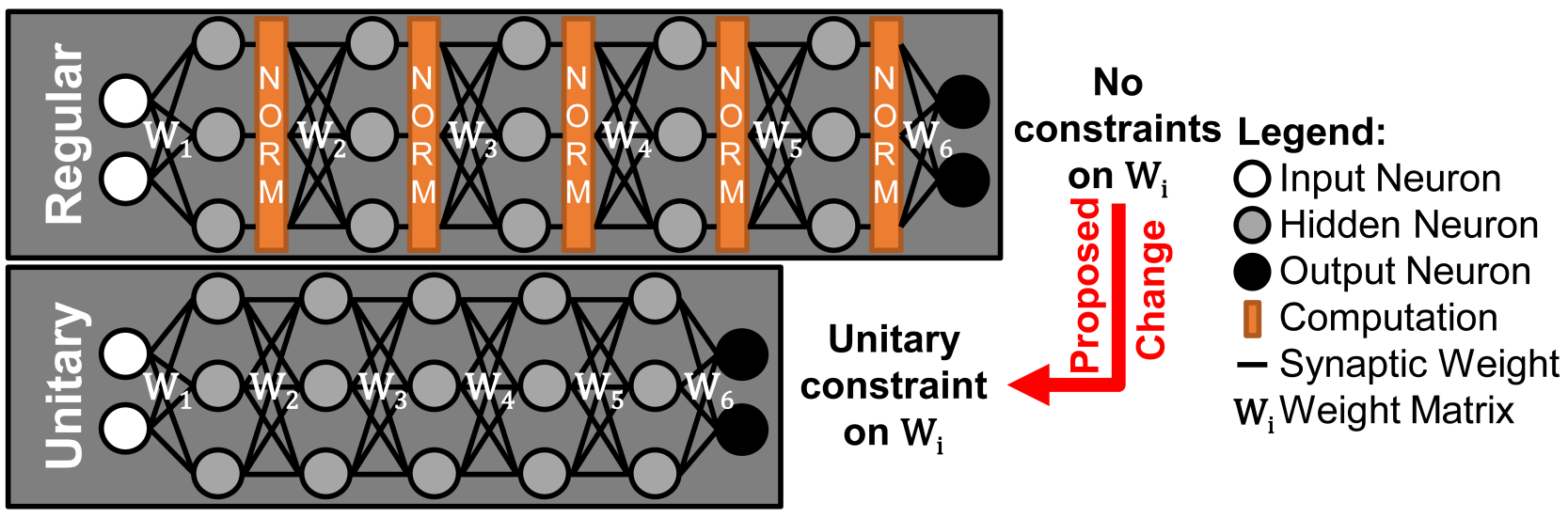

**Fig. 1.** Unitary network for mitigating exploding and vanishing activations.

[1] Unitary matrices can have complex values. When the matrices only contain real components, they are called orthogonal matrices, which is a subset of unitary matrices, and our proposal works in both cases. The eigenvalues of a unitary matrix have modulus 1.

### 1.3 Literature Review

Lie algebra is not the only way to construct unitary matrices. Researchers have explored many options to construct unitary weights for RNNs, including eigendecomposition [13], Cayley transform [14], square decomposition [15], Householder reflection [16], and optimization over Stiefel manifolds [17]. These methods decompose the unitary matrix into smaller parameter spaces with mathematical processes that guarantee unitarity; however, the weight matrices in these approaches must be square. For convolutional neural nets with rectangular weights, there are approximation techniques based on least square fitting [18], singular value decomposition [19], and soft regularization [20] due to the additional complexity of rectangular filters. These techniques find the best approximates to the unitary weights, but they do not guarantee the weight matrices to be strictly unitary. *On the contrary, our approach combines the best of the two schools—it is both strictly unitary and applicable to non-square matrices.* Our work is the first report of applying the unitary weights based on the Lie algebra framework for a deep convolutional neural network with a comprehensive performance study, aiming to make the unitary network an attractive alternative to conventional normalization methods in inference-time-critical applications.

## 2 Unitary Neural Networks with Lie Algebra

### 2.1 Square Unitary Weight Matrices

In this section, we explain the mathematical framework [12] for representing the unitary group with orthogonal matrices, collectively known as the Lie group [21]. Linearization of the Lie group about its identity generates a new set of operators; these new operators form a Lie algebra. Lie algebra is parameterized by the Lie parameters, which we arrange as a traceless lower triangular matrix, $\boldsymbol{L}$. We name it Lie parameters because it contains independent trainable parameters for the neural networks. The representable algebra through this parameterization is only a subspace of unitary groups, and it is sufficient for guaranteeing signal stability in deep neural networks.

The Lie parameters ($\boldsymbol{L}$) are related to the Lie algebra ($\boldsymbol{A}$) by the following equation:

$$\boldsymbol{A} = \boldsymbol{L} - \boldsymbol{L}^T, \tag{1}$$

where $T$ corresponds to taking the matrix transpose. An essential feature of matrix $\boldsymbol{A}$ is that it is a skew-symmetric matrix, i.e., $\boldsymbol{A}^T = -\boldsymbol{A}$, because any compact metric-preserving group, including the orthogonal group, has anti-symmetric Lie algebra [22]. Furthermore, the following equation proves that the chosen representation for the Lie parameters will produce an anti-symmetric Lie algebra:

$$\boldsymbol{A}^T + \boldsymbol{A} = (\boldsymbol{L} - \boldsymbol{L}^T)^T + \boldsymbol{L} - \boldsymbol{L}^T = 0. \tag{2}$$

Additionally, in the last step of our pipeline to construct unitary matrices, we exponentiate the Lie algebra, $\boldsymbol{A}$, to obtain the group representation, which will be a unitary matrix ($\boldsymbol{U}$):

$$\boldsymbol{U} = EXP(\boldsymbol{A}) = \sum_{N=0}^{\infty} \boldsymbol{A}^N / N!. \quad (3)$$

We approximate this matrix exponentiation with an 18-term Taylor series in our implementation. Besides eliminating any term beyond the 18[th] order in Eq. (3), we efficiently group the computation to avoid redundant multiplications, a standard approach used in many matrix computation software to save time [23, 24].

Suppose the neural network has square weight matrices. In that case, we can use the unitary matrices ($\boldsymbol{U}$) to replace the original weights, forcing the neural signals to maintain their norms without explicit normalization. We can train the Lie parameters using backpropagation and automatic differentiation because all steps in the pipeline above are algebraic functions [25, 26]. As mentioned previously, researchers have only applied the unitary pipeline to a small recurrent neural network (RNN), which has a single *square* weight matrix repeatedly applied in time [12]. Nevertheless, the requirement for the weights to be square severely limits the usefulness of the presented framework. Using the Lie algebra formalism to construct unitary weights is an elegant method to regulate signals, and we wish to find a way to bring this concept to deep convolutional neural nets with any non-square weight matrices.

## 2.2 Unitary Weight Matrices of Any Shapes and Dimensions

In the above section, the weight matrices must be square, forcing the number of neurons for both the input and output of a particular layer to be identical. This requirement cannot be satisfied in most convolutional neural nets. Convolutional layers have weight matrices commonly referred to as "filters." These filters will convolute with the input image as the following [5]:

$$\boldsymbol{O}(i,j) = (\boldsymbol{I} * \boldsymbol{F})(i,j) = \sum_m \sum_n \boldsymbol{I}(i+m, j+n)\boldsymbol{F}(m,n), \quad (4)$$

where $\boldsymbol{O}$ is the output activation map of this layer, $\boldsymbol{I}$ is the input image, and $\boldsymbol{F}$ is the convolutional filer. An example of convolution is illustrated in Fig. 2(a).

Moreover, we can succinctly represent the convolution as a single dot product through the Toeplitz matrix arrangement [27, 28]. Suppose we arrange the input image as a Toeplitz matrix and flatten out the filters to a 2-dimensional weight matrix. In that case, the convolution simplifies to a dot product between the Toeplitz matrix of the image and the flattened filter weights. Effectively, we convert the convolution between high-dimensional tensors to *multiplications* between 2-dimensional matrices. These flattened filters are usually rectangular $m$ x $k$ matrices, where $m \neq k$. If $m$ matches $k$, the weight matrix is square, allowing us to apply the unitary pipeline to ensure each row of the Toeplitz matrix will maintain its norm. On the other hand, when dealing with rectangular weights, we need to handle them with special care to achieve the desired effect of norm preservation.

Our innovation is that we discard the excess columns in the unitary matrix when $m \neq k$ (i.e., when the weight matrix has unequal width vs. height); for now, we will assume $m > k$ because the other cases only require a few slight adjustments. Even though there is no way around the fact the unitary matrices must be square, *we discovered that it is unnecessary to use the whole unitary matrix: we can just take the* *first few columns that we need.* We will construct the unitary matrix as a $m$ x $m$ matrix (i.e., in the larger of the two dimensions). This way, we can reuse the existing pipeline in Sect. 2.1. Our proposed pipeline is as follows. We only keep the first $k$ columns of the Lie parameters, setting everything else to zero. Likewise, we only take the first $k$ columns of the resulting unitary matrix (***U***), discarding the rest (Fig. 2(b)). Because the rectangular matrix now has the correct dimensions, we multiply the input image (in the Toeplitz form) with the unitary weight matrix. Below is a summary of our process to construct the unitary rectangular weights in the mathematical form:

$$\bar{y} = \boldsymbol{W}\bar{x}, \tag{5}$$

where

$$\boldsymbol{W} = [\,\overline{u_1} \quad \overline{u_2} \quad \dots \quad \overline{u_k}\,] \in R^{m \times k}. \tag{6}$$

$\overline{u_1} \dots \overline{u_k}$ are the first $k$ column vectors from the unitary matrix $\boldsymbol{U}$ (Note that when $m \leq k$, we still construct the unitary matrix $\boldsymbol{U}$ in the larger dimension, but $\boldsymbol{W}$ will be transposed to obtain the desired dimensionality for the matrix multiplication). Lastly, the output vector is explicitly normalized using the Euclidean metric when $m > k$; this step is not required for $m \leq k$:

$$\overline{y_{final}} = \begin{cases} \bar{y}/\|\bar{y}\|_2 \; for \; m > k \\ \bar{y} \; for \; m \leq k \end{cases}, \tag{7}$$

where $\|\cdot\|_2$ denotes the Euclidean norm (a.k.a., Euclidean metric or the L2 norm), a distance measure calculated by squaring all the coordinates, summing the results, and taking the square root.

In theory, it is possible to avoid the explicit normalization Eq. (7) completely by one of the following two ways: by partitioning the tall rectangular weight matrix into a vertical stack of smaller matrices that are either square or wide. Or, by exploring alternative mappings from the various dimensions of $\boldsymbol{W}$ to $m$ and $k$ to ensure $m \leq k$. Nevertheless, we took the direct normalization approach in this work for conceptual clarity, and it is only required in a small portion of the network. Moreover, even though it is not ideal to add normalization back to portions of our network, the unitary weights offer other benefits over conventional normalization. Researchers have found orthogonal weights lead to more efficient filters with fewer redundancies [20]. Our normalization process does not add additional training parameters to keep track of the activations' mean and variance.

Discarding columns of unitary matrices has important geometrical meanings. A unitary matrix represents a rotation in the n-dimensional space when $m = k$; additionally, its columns form a complete set of orthonormal bases in the rotated coordinate system. We have utilized the latter to paint a geometric understanding of our procedure—each of the $k$ columns is an orthonormal basis in the m-dimensional space. For $m > k$, the unitary weight ($\boldsymbol{W}$) is projecting an input row vector $\bar{x}$ to a lower-dimensional manifold spanned by the unitary matrix's first $k$ columns, a subset of orthonormal bases. When we multiply the Toeplitz matrix with this unitary weight matrix, we perform a dot product between the row vectors against each orthonormal basis, measuring how much the input vector aligns with a specific basis. According to the Pythagorean theorem, this projection will produce a shorter vector than the original one because we dispose of those vector components associated with the unitary matrix's discarded columns. As a result, we need to normalize the output to recuperate the signals lost in missing dimensions.

On the contrary, for $m < k$, each row of the weight matrix is an orthonormal basis. In that case, we are mixing the orthonormal bases according to the ratio prescribed by the input row vector $\bar{x}$, resulting in a higher-dimensional output vector $\bar{y}$. This dimensionality expansion happens when we multiply the weight matrix with a row vector ($\bar{x}$) of the Toeplitz matrix that encodes the input image. Effectively, we are projecting a vector to the larger dimensions through the wide unitary weight ($\boldsymbol{W}$), and this operation preserves the Euclidean norm of the input vector $\bar{x}$. To prove this property mathematically, we simply compare the norm of $\bar{x}$ against the norm of $\bar{y}$. When we use the orthonormal bases defined by the unitary matrix ($\boldsymbol{U}$) to describe vector locations, the first p coordinates of $\bar{y}$ match $\bar{x}$, and the rest of the coordinates are zeros. Hence, $\|\bar{y}\|_2$ is the same as $\|\bar{x}\|_2$ because the Euclidean norm is defined as the square root of the sum of the squared coordinates.

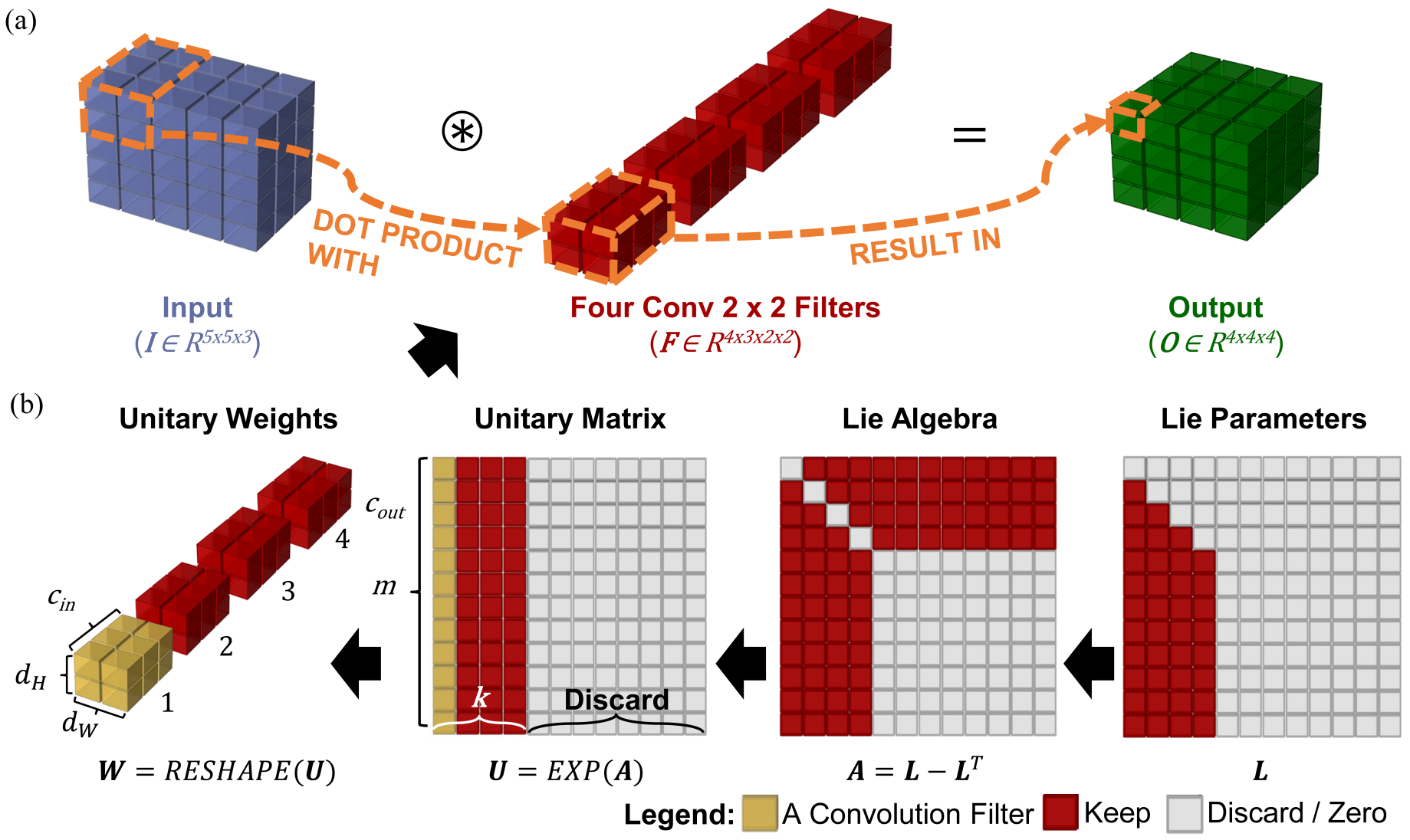

**Fig. 2.** Pipeline to construct unitary weights for a convolutional neural network. (a) convolution between an example input image and filters. (b) our proposed way of constructing unitary weights of any dimensions. From the right, Lie parameters contain all the trainable parameters, and we only need the first $k$ columns. Similarly, we keep the first $k$ columns of the resulting unitary matrix and reshape them to match the desired dimensionality for the convolutional filters. In this example, $k = c_{out}$ and $m = c_{in} \times d_H \times d_W$. This mapping will depend on the target applications.

# 3 Experiments

## 3.1 Network Architecture

We applied the proposed unitary weight matrices to the residual neural network (ResNet) for image recognition; our architecture is a narrower and shorter variant of the popular ResNet-50 [3]. We picked a smaller model to prevent overfitting to the training data because ResNet-50 was designed for the more complex ImageNet dataset. Our network (uResNet-44) has only 43 convolutional layers with a fully connected layer at the end for projecting the high-dimensional neural signals to ten output classes. We documented the sizes and number of convolutional filters in Fig. 3 for reproducibility. Also, we studied the scalability in terms of depth with the 92 and 143-layer networks (uResNet-92 and uResNet-143). See our source code for details[2].

## 3.2 Dataset

We used the CIFAR-10 image recognition dataset created by the Canadian Institute for Advanced Research, and it contains 60,000 32 x 32 color images with ten labeled classes. The recognition task is to predict the correct class of each image in the dataset. CIFAR-10 is freely available for download [29]. We split the dataset into 50,000 training and 10,000 test images with the same data argumentation scheme as the original ResNet paper [3]. We also tested our unitary neural network's susceptibility to overfitting with the CIFAR-100 dataset [29].

## 3.3 Training Details

We modified the source code found in this reference [30] for comparison against conventional normalization techniques, sharing the same learning rate (0.1), learning schedule (divide by 10 at 100, 150, and 200 epochs), batch size (128), and training epochs (250). The only modification we made for the unitary neural net is that we added the unitary pipeline using the method described previously in Sect. 2.2 for the convolutional layers. We also removed all the normalization blocks in the unitary version. We trained the regular and the unitary networks with the stochastic gradient descent optimizer in PyTorch with a momentum setting of 0.9 and weight decay of 2e-4. We measured the neural networks' speed and memory usage by simulating each neural architecture one at a time on a single NVIDIA RTX3090 graphics card with 24 GB total video memory.

## 3.4 Caching of the Unitary Weights

During training, the entire neural pathway is enabled, including the block that contains the Lie parameters, Lie algebra, and Lie group (Fig. 3). Gradients are backpropagated from the output to update the Lie parameters. After training is complete, the best unitary weights are cached; thus, we do not need to recompute the unitary weights during inference.

[2] https://github.com/h-chang/uResNet

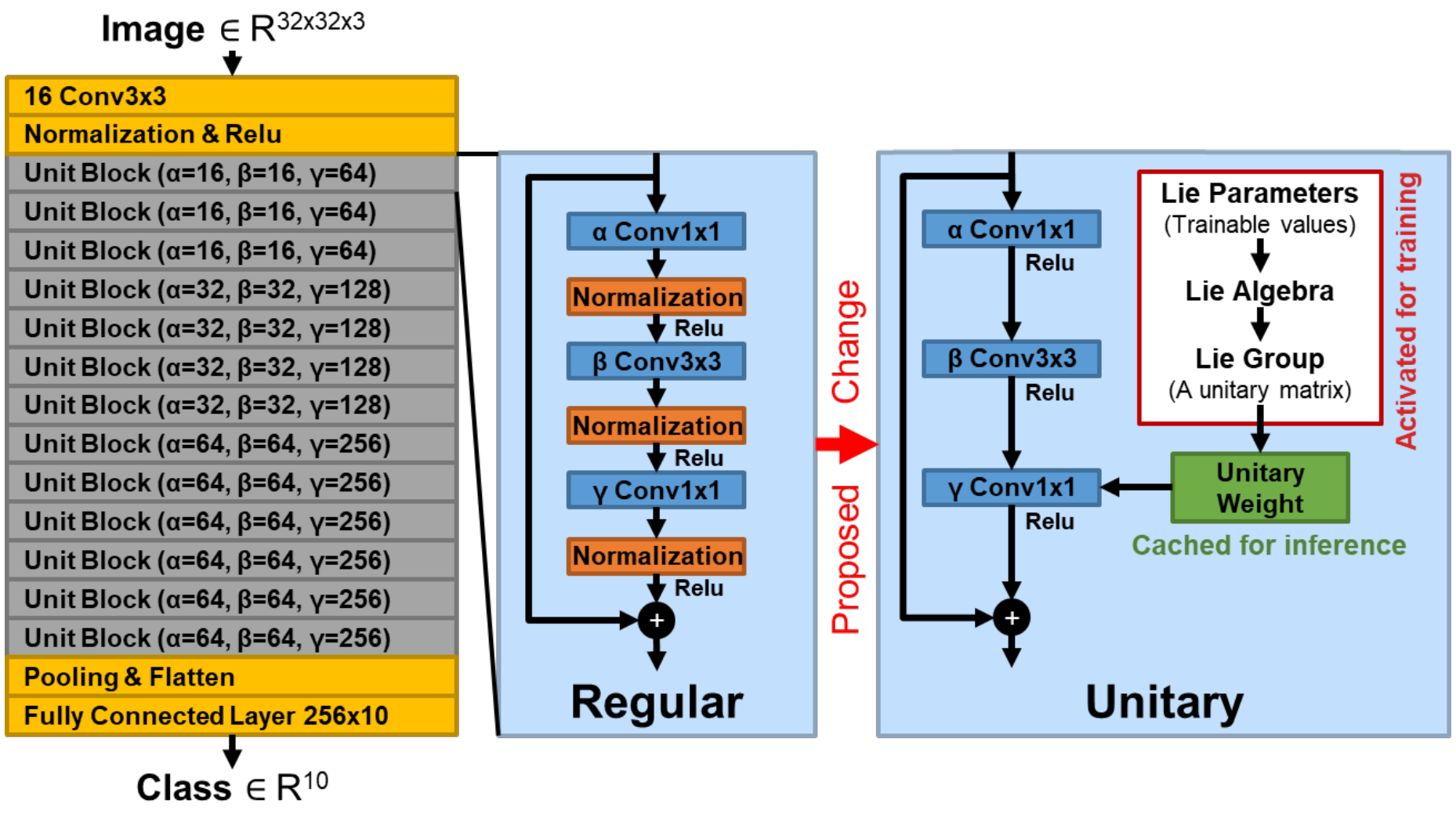

**Fig. 3.** The unitary convolutional neural network (CNN) architecture. There are two main differences between the regular and the unitary CNN. Firstly, in unitary CNNs, we permanently remove the normalization blocks to speed up computation because the unitary weights already preserve the signal strengths across layers. Secondly, unitary CNNs have an additional Lie block labeled "Activated for training" on the right. The Lie block is active only during the training mode to learn the Lie parameters. At the end of the training mode, a set of unitary weights is constructed from the Lie parameters and cached. The convolutional filters will use these pre-recorded unitary weights during inference. The removal of normalization significantly improves inference speed. We enlarge one of the unit blocks to illustrate its content; each unit block contains three convolutional layers. Unit blocks are cascaded to create a feed-forward convolutional neural network. The only difference between the unit blocks is the number of convolutional filters, which we label as $\alpha$, $\beta$, and $\gamma$ in the figure. $\alpha$ Conv1x1 for a layer with $\alpha$ = 16 means that there are 16 convolutional filters with size 3 x 3 in that layer. The rectified linear unit (Relu) is used as nonlinearity at locations depicted in the figure.

## 4 Results and Discussion

With our proposed unitary convolutional neural network from Sect. 3, we compare the performance of our proposal against popular normalization methods and summarize the main results of our experiment in Fig. 4 below. By removing the network's unitary pipeline (the block with Lie parameters, Lie algebra, and Lie group in Fig. 3) during test time, we achieved a much faster inference speed than other normalization methods, including the batch norm [10], group norm [31], layer norm [32], and instance norm [33]. Each of these methods addresses a specific problem; therefore, the designer might favor one over the other depending on the application. *With our uni-* *tary convolution, we offer the community another tool in the toolbox that is lightning fast—32% faster than the instance norm in inference.* We compute the speedup by dividing the inference time of the unitary norm with the instance norm's in Fig. 4(e). Our method shares many characteristics with the instance norm; however, instead of normalizing based on the neural signals' statistics, we devise a set of unitary weights to ensure signals maintain their norm per Toeplitz matrix row. Compared to the instance norm's training time, our training time for the unitary network is also long due to the need to perform matrix exponentiation. Still, it is possible to further expedite it by limiting the frequency that we exponentiate (i.e., sharing the same unitary weights for several iterations). The result shown in Fig. 4 is measured without weight sharing during training; we will report further improvements in the future.

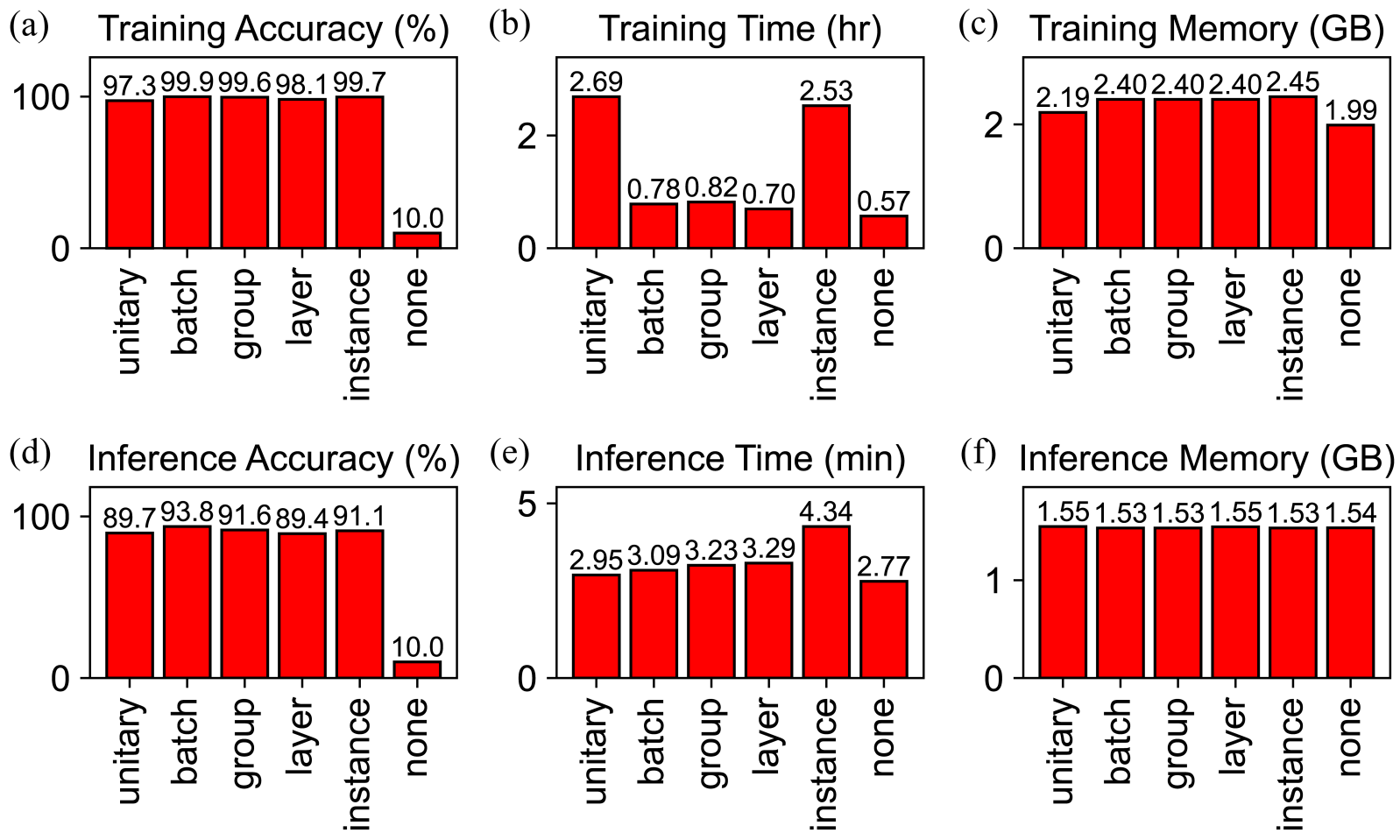

**Fig. 4.** Performance comparison between our proposed unitary convolutional network and other normalization methods. Other methods include the batch norm [10], group norm [31], layer norm [32], and instance norm [33]. We also included the case without any normalization for comparison. All metrics are average measurements over four simulation runs. We used the residual network (ResNet) architecture with 43 convolutional layers to measure the accuracy, the time, and the memory benchmarks when the networks perform image recognition tasks on the CIFAR-10 dataset, which has ten unique classes of objects. The accuracy reports the percentage of time the network determines the image class correctly with one try. (a, d for training, inference accuracies, respectively). The training time is the time to train the network with 12.5 million images, while the inference time reports the time to recognize 2.5 million images (b, e). Because we trained these networks on graphics processors, memory benchmarks measure the maximum video memory a network consumed during each operation mode (c, f).

In our experiment, both unitary network and batch normalization do not calculate running statistics (i.e., means and variances) during inference while group, layer, and instance norms track running statistics in the test set. Batch normalization is the second fastest and can potentially match the speed of the unitary network if the batch normalization layer is absorbed into the previous convolutional filters. However, batch normalization will not perform well in applications that require small batch sizes or normalization per data sample such as making adjustments to the contrast of individual images [33]. Group, layer, and instance normalizations work on a per-image basis; the difference between them is the number of channels that they average over. In our experiment, we picked a group size of eight; hence, the group normalization needs to keep track of eight means and variances per image. Contrary to layer norm that only requires one mean and one variance per image, instance norm track as many means and variances as the number of channels, which is up to 256 in our architecture. Our unitary network maintains the L2 norm of each row in the Toeplitz matrix representation per image, delivering similar effects as the instance norm but without the inference speed penalty. The mapping between the filters and the unitary matrix determines which dimension of the activation map that the unitary network is effectively normalizing. For this reason, practitioners should assign $c_{out}$, $c_{in}$, $d_H$, and $d_W$ to $m$ and $k$ in Fig. 2 differently based on the target applications.

The unitary network also uses less temporary memory (dynamic random-access memory or DRAM) required to backpropagate neural signals through the normalization layers during training; more specifically, 8% less than all other normalization methods. Despite our advantages in inference speed and training memory, unitary networks' accuracy is slightly lower in general. Unitary weights constrain the signals to be on the n-sphere (or k-sphere since we have $k$ dimensions) by design and are less expressive than free weights. Nevertheless, our accuracy is comparable to other normalizations and even surpasses the inference accuracy of layer norm. An additional advantage for the unitary network is apparent when we save the model parameters to hard disks: as we demonstrated in Fig. 2, the matrices encoding the Lie parameters have many zeros, which lead to better compression of the parameter files. An approximation for model size saving is roughly a 15% to 50% reduction in disk space when working with unitary convolutional architectures. We compute the 50% reduction by leveraging the fact that we only need to record half of the values in a triangular Lie parameter matrix, assuming that the weight matrix is square.

Our unitary neural networks are less susceptible to overfitting. Using the CIFAR-100 dataset and the same network structure (uResNet-44), we discovered that unitary networks have a smaller gap between the training loss (1.44) and the testing loss (1.62). While Regular neural networks with batch normalization have a larger gap between the training loss (0.0699) and the testing loss (1.56). Furthermore, our unitary networks can be deepened to 100+ layers without the costly normalization blocks: the 92-layer version (uResNet-92) achieves 99.6% and 90.4% in training and testing accuracies, respectively, on CIFAR-10. Similarly, the 143-layer version (uResNet-143) delivers 99.7% and 90.7% in training and testing accuracies.

## 5 Conclusion

We report here the first instance of using unitary matrices constructed according to the Lie algebra for rectangular convolutional filters, which eliminates the exploding and vanishing activations in deep convolutional neural networks. With clear geomet-rical interpretations, our theory is a breakthrough based on rigorous, exact construction of the unitary weights applicable to all types of neural networks including but not limited to convolution. The key innovation is that we found a way to ensure signal unitarity with unitary weight matrices of any shapes and dimensions such that the neural signals will propagate across the network without amplification or degradation. Moreover, unlike traditional normalization, our approach has the least impact on inference time, achieving a 32% speedup in recognizing color images when compared to instance normalization. The effective normalization dimension is adjustable in our framework through the mapping between the convolutional filters and the unitary matrices. Our proposal also reduces hard disk storage by up to 50% depending on the neural architectures. The presented framework establishes unitary matrices as a design principle for building fundamentally stable neural systems.